\documentclass[pdflatex,sn-mathphys]{sn-jnl}

\jyear{2021}%

\theoremstyle{thmstyleone}%
\theoremstyle{thmstyletwo}%
\theoremstyle{thmstylethree}%

\usepackage{subfigure}
\begin{document}

\title[Global-Local Dynamic Feature Alignment Network for Person Re-Identification]{Global-Local Dynamic Feature Alignment Network for Person Re-Identification}


\author{\fnm{Zhangqiang} \sur{Ming}}\email{mingzhangqiang@stu.scu.edu.cn}

\author{\fnm{Yong} \sur{Yang}}
\author{\fnm{Xiaoyong} \sur{Wei}}
\author{\fnm{Jianrong} \sur{Yan}}
\author{\fnm{Xiangkun} \sur{Wang}}

\author{\fnm{Fengjie} \sur{Wang}}

\author*[]{\fnm{Min} \sur{Zhu*}}\email{zhumin@scu.edu.cn}


\affil{\orgdiv{College of Computer Science}, \orgname{Sichuan University}, \orgaddress{\city{Chengdu}, \postcode{610065}, \country{China}}}




\abstract{The misalignment of human images caused by bounding box detection errors or partial occlusions is one of the main challenges in person Re-Identification (Re-ID) tasks. Previous local-based methods mainly focus on learning local features in predefined semantic regions of pedestrians. These methods usually use local hard alignment methods or introduce auxiliary information such as key human pose points to match local features, which are often not applicable when large scene differences are encountered. To solve these problems, we propose a simple and efficient Local Sliding Alignment (LSA) strategy to dynamically align the local features of two images by setting a sliding window on the local stripes of the pedestrian. LSA can effectively suppress spatial misalignment and does not need to introduce extra supervision information. Then, we design a Global-Local Dynamic Feature Alignment Network (GLDFA-Net) framework, which contains both global and local branches. We introduce LSA into the local branch of GLDFA-Net to guide the computation of distance metrics, which can further improve the accuracy of the testing phase. Evaluation experiments on several mainstream evaluation datasets including Market-1501, DukeMTMC-reID, CUHK03 and MSMT17 show that our method has competitive accuracy over the several state-of-the-art person Re-ID methods. Specifically, it achieves 86.1\% mAP and 94.8\% Rank-1 accuracy on Market1501.}




\maketitle
\section{Introduction}\label{Introduction} 
\begin{figure}[tbh]
	\centering
	\includegraphics[scale=0.58]{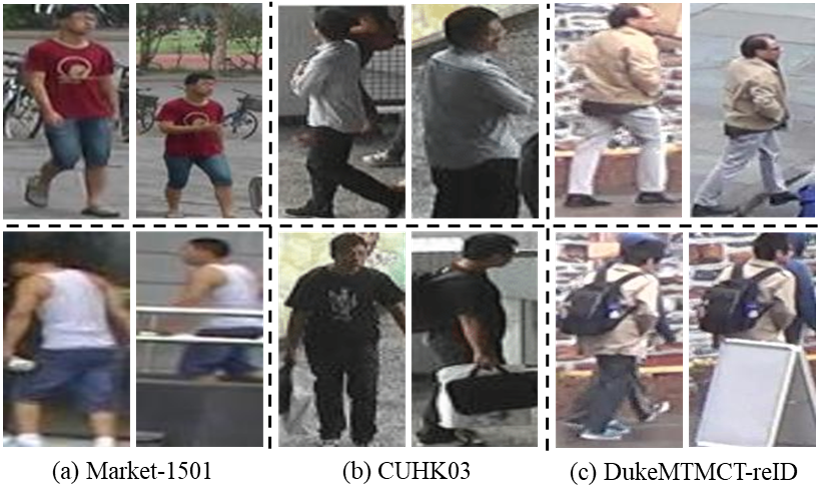}
	\caption{Examples of bounding box detection errors (first row) and partially occluded (second row) images from Market1501 \cite{Zheng2015Scalable}, CUHK03\cite{Wei2013HumanReidentification} and DukeMTMC-reID\cite{Zheng2017Unlabeled}, respectively.} 
	\label{fig:1-1-Challenge}
\end{figure}

Person Re-Identification (Re-ID) is a challenging task in the field of computer vision, aiming to determine whether a person captured by different cameras or person images from different video clips of the same camera is the same person. However, due to the complexity of realistic scenarios, person Re-ID still faces many challenges, like person  bounding box detection errors and occlusion, as shown in \autoref{fig:1-1-Challenge}, making it difficult for people to identify a specific person from large gallery collections. To address these challenges, most prior works focused on learning global features of pedestrians using Convolutional Neural Networks (CNN), the idea of which can be summarized mainly as representational learning and deep metric learning\cite{LUO201953AlignedReID}. Traditional representation learning methods aimed to learn the rigid and invariant features \cite{Zheng2016Person,Zheng2017Discriminatively,Zheng2017Wild,Chen2018DeepTransfer}, and most deep metric learning methods aimed to reduce the distance of the same person \cite{Ristani2018Multitarget,Shi2016Embedding,Hermans2017InDefense,Chen2017BeyondTriplet}. However, all these methods learn features from the entire image and contain only the coarse-grained global information of the pedestrian, while ignoring local key details. 

To better learn local features, some methods \cite{Sun2018BeyondPart,Sun2019Perceive,Varior2016Siamese,Fu2018Horizontal} use horizontal stripes or grids to extract local features of person body parts, but such methods require pre-adjust pedestrian alignment to obtain good performance. Some researchers \cite{Liu2017HydraPlus,Chen2019ABDNet,Li2018Harmonious,Chen2019Mixed} also present attentional mechanisms to complement discriminative features while bringing in extra background attention, which affects the final feature representation of the person. There are also some works \cite{Zhao2017Spindle,Zhao2017DeeplyLearned,Su2017PoseDriven,2020MultiFang,2020FineZhou,2020SemanticsJin} using human pose points obtained by human pose estimation models to match different body parts or align key pose points, but training such models requires a huge amount of label data, and additional computational resources are consumed. Alternatively, some researchers \cite{LUO201953AlignedReID,Su2017PoseDriven,Wei2019GLAD,Wang2018LearningDiscriminative,Zheng2019Pyramidal,2020SemanticsJin,2021HOReIDWang,2021PersonLi} combine global and local features to enhance the final pedestrian diversified feature representation. In general, the pose-guided alignment methods require additional computational resources. The local hard alignment methods are hard to obtain high accuracy of person Re-ID when encountering large scene differences such as  bounding box detection errors and partial occlusions.

In this paper, we propose an efficient Local Sliding Alignment (LSA) strategy. We divide the original image into horizontal stripes, and the local features are more focused on finer discriminative features in each strip. LSA achieves dynamic alignment of local stripes by setting sliding windows for local stripes of pedestrians and calculating the shortest alignment distance of local stripes within each sliding window. Compared with the methods of local hard alignment and pose guided alignment, LSA can effectively alleviate the problem of unaligned images due to bounding box detection errors and partial occlusions. It does not require the introduction of additional supervision information. To explore global and local information, we design a Global-Local Dynamic Feature Alignment Network (GLDFA-Net) framework for the person Re-ID task, which contains both global and local branches that learn feature representations at different granularities. We introduce LSA into the local branch of GLDFA-Net to guide the computation of distance metrics. LSA is not only designed to be simple and efficient but also can effectively improve the accuracy of the model. 

More specifically, in the training phase, we use LSA to calculate the local alignment distances and combine them with the global distances, together as the sample distance for triplet loss. LSA can shorten the distance of samples with the same ID, thus effectively mining hard samples and guiding the model to learn more discriminative features. In the inference stage, we use LSA to calculate the local alignment distance for similarity measurement, which can further improve the accuracy of person matching. Compared with previous local-based methods, our method is an end-to-end learning process with a simple and efficient network structure, requires only dividing horizontal stripes as local features. Experiments show that our method performs better on multiple mainstream Re-ID datasets.

In summary, the main contributions of this article are summarized as follows:
\begin{itemize} 
	\itemsep 0.5em 
	\item We propose a new method named Local Sliding Alignment (LSA) that achieves dynamic alignment of local features by setting sliding windows for their local stripes and calculating the shortest alignment distance of the stripes within the corresponding sliding windows. LSA can effectively suppress spatially misalignment and the noises from unshared regions. It does not require the additional auxiliary pose information.
	
	\item We design a Global-Local Dynamic Feature Alignment Network (GLDFA-Net) framework, which contains two branches, global and local. We introduce LSA into the local branch of GLDFA-Net to guide the computation of the distance metric, which can further improve the accuracy of the testing phase.
	
	\item The experiments show that our method achieves significant improvements in Rank-1 accuracy and mAP on the Market1501 \cite{Zheng2015Scalable}, DukeMTMC-reID \cite{Zheng2017Unlabeled}, CUHK03 \cite{Wei2013HumanReidentification} and MSMT17 \cite{Wei2018PersonTransfer}.
\end{itemize} 
\section{Related Works}\label{RelatedWorks}
In this section, we review some work that is closely related to this paper, including person Re-ID methods based on deep feature learning and based on local feature alignment.
\subsection{Deep Feature Learning based Person Re-ID}
In recent years, with the rapid development of deep learning, deep learning techniques  have been widely used in person Re-ID tasks, and have achieved high retrieval accuracy \cite{Zheng2016Person,Zheng2017Discriminatively,Ristani2018Multitarget,Shi2016Embedding}. Compared with traditional methods, deep neural networks can automatically learn global discriminative features, and only a simple metric function is needed to determine the similarity of two pedestrian images. Current deep learning-based person Re-ID methods can be categorized into two types \cite{LUO201953AlignedReID,Wei2019GLAD}, i.e., representation learning methods and distance metric learning methods. Representation learning methods \cite{Zheng2016Person,Zheng2017Discriminatively,Zheng2017Wild,Chen2018DeepTransfer,Wang2018LearningDiscriminative} aim to learn the powerful and distinguishing features of pedestrian images. Zheng et al. \cite{Zheng2016Person,Zheng2017Discriminatively} regard the training process of person Re-ID as a multi-classification problem of images and proposed an ID Discriminative Embedding (IDE) network model, which treats each pedestrian as a different class and uses the pedestrian's ID as a classification label to train the deep neural network. Zheng et al. \cite{Zheng2017Discriminatively} and Chen et al. \cite{Chen2018DeepTransfer} jointly verification loss and ID loss can lead to better performance of CNN.

Different from the representation learning method, distance metric learning methods \cite{Ristani2018Multitarget,Shi2016Embedding,Hermans2017InDefense,Chen2017BeyondTriplet,Mishchuk217WorkingHard,Cheng2016MultiChannel,Luo2019Tricks,Xiao2017Margin} aim to learn a mapping from the original image to the feature embedding. This method makes the distance in the feature space smaller for the same pedestrian and larger for different people \cite{Song2016DeepMetricLearning}. Triplet loss is the most common distance metric learning method. Its principle is to define two images of the same person as a positive pair, while two images of different people are defined as a negative pair to guide the network model to training \cite{Ye2020Deep,2021DeepMing}. The traditional triplet loss may have simple sample combinations and a lack of training for hard sample combinations. For this reason, some researchers consider improving the triplet loss for mining hard samples \cite{Ristani2018Multitarget,Hermans2017InDefense,Xiao2017Margin}. Luo et al. \cite{Luo2019Tricks} jointly trained the classification (ID) loss and the metric loss to accelerate the convergence of the model. However, the traditional triplet loss has a poor ability to constrain the samples within the class. The center loss can shorten the distance of the samples within the class, which can effectively make up for this defect of triplet loss. Therefore, we use classification loss, triplet loss with hard sample mining and center loss joint constraint model in our proposed method.

\subsection{Local Features Alignments based Person Re-ID} 

Traditional global feature learning methods mainly focus on the appearance and spatial location information but lack the learning of fine-grained local information. Recent works \cite{Sun2018BeyondPart,Sun2019Perceive,Varior2016Siamese,Fu2018Horizontal,Liu2017HydraPlus,Chen2019Mixed,Li2018Harmonious,Chen2019ABDNet,Zhao2017Spindle,Zhao2017DeeplyLearned,Su2017PoseDriven,Miao2019PoseGuided,Kalayeh2018HumanSemantic,2021HOReIDWang} further improve the accuracy by learning deep local features. We have summarized three main approaches based on local feature learning: determining body parts according to predefined partitions, locating body parts through the joint point method, and locating local features through spatial attention. Some stripe-based methods \cite{Varior2016Siamese,Fu2018Horizontal,Sun2018BeyondPart,Sun2019Perceive} divide the deep feature maps into regions according to certain predefined division rules to learn the local features.
Li et al. \cite{2021PersonLi} propose a Re-ID model with Part Prediction Alignment (PPA), which aims at aligning the predicted distributions between each part.

Ngo et al. \cite{2005MotionNgo,2008BeyondNgo} explored the method of high-level feature extraction and aim to explore context-based concept fusion by modeling inter-concept relationships, which are modeled not based on semantic reasoning. Domain adaptation aims to learn an adaptive classifier for target data using the labeled source data from a different distribution. Some methods \cite{2021HazyPang,2021UnifiedLiu,2021MetaJi} experiment with the adaption of various person Re-ID domains through knowledge transfer to improve the generalization of the person Re-ID model. Semantic alignment-based person Re-ID has been investigated in some works \cite{Zhao2017Spindle,Zhao2017DeeplyLearned,Su2017PoseDriven,2020FineZhou,2020MultiFang,2020SemanticsJin,2021ProceedingsYang}. Some methods locate the key points of body parts to extract the body regions and then use CNN to capture semantic features from different body regions to obtain discriminative local details information. In addition, some recent studies \cite{Liu2017HydraPlus,Chen2019Mixed,Li2018Harmonious,Chen2019ABDNet,2020HeterogeneousYang,2021SelfZhang} have introduced attention as a supplement to the discriminative features and achieved good performance.

However, local feature learning methods tend to learn detailed information about a certain region of the pedestrian, but their reliability may be affected by detection of bounding box errors and partial occlusions. In addition, most methods only focus on the parts with fixed semantics, cannot cover all the distinguishing information. Therefore, researchers often combine the methods of local and global feature learning \cite{Su2017PoseDriven,Wei2019GLAD,Wang2018LearningDiscriminative,Zheng2019Pyramidal,Chen2020SalienceGuided,Zhang2020RelationAware,Yao2019DeepRepresentation} to enhance the final distinguishing feature representation. Wang et al. \cite{Wang2018LearningDiscriminative} proposed a feature learning strategy that combines global and local information and designed a Multi-Granularity Network (MGN) to learn features of different granularities from pedestrian images. Wei et al. \cite{Wei2019GLAD} proposed the Global-Local-Alignment Descriptor (GLAD), which learns the global and three local features separately and finally connects them to form a distinctive and robust GLAD.

In general, the above methods either use hard alignment to match local features or introduce additional supervision to aid local alignment, with little attention to the dynamic alignment of local features. In this paper, we combine the global features of pedestrians and the local features of horizontal stripe partitions and use a simple and efficient local sliding alignment strategy to dynamically align discriminative features of pedestrians.

\section{Methods}\label{Methods}
In this section, we first describe the network structure of Global-Local Dynamic Feature Alignment Network (GLDFA-Net), then introduce the Local Sliding Alignment (LSA) strategy. Finally, we design the loss function of the model proposed in this paper.
\begin{figure*}[t]
	\centering	
	\begin{minipage}[t]{0.95\linewidth}
		\centering
		\includegraphics[width=1.0\linewidth]{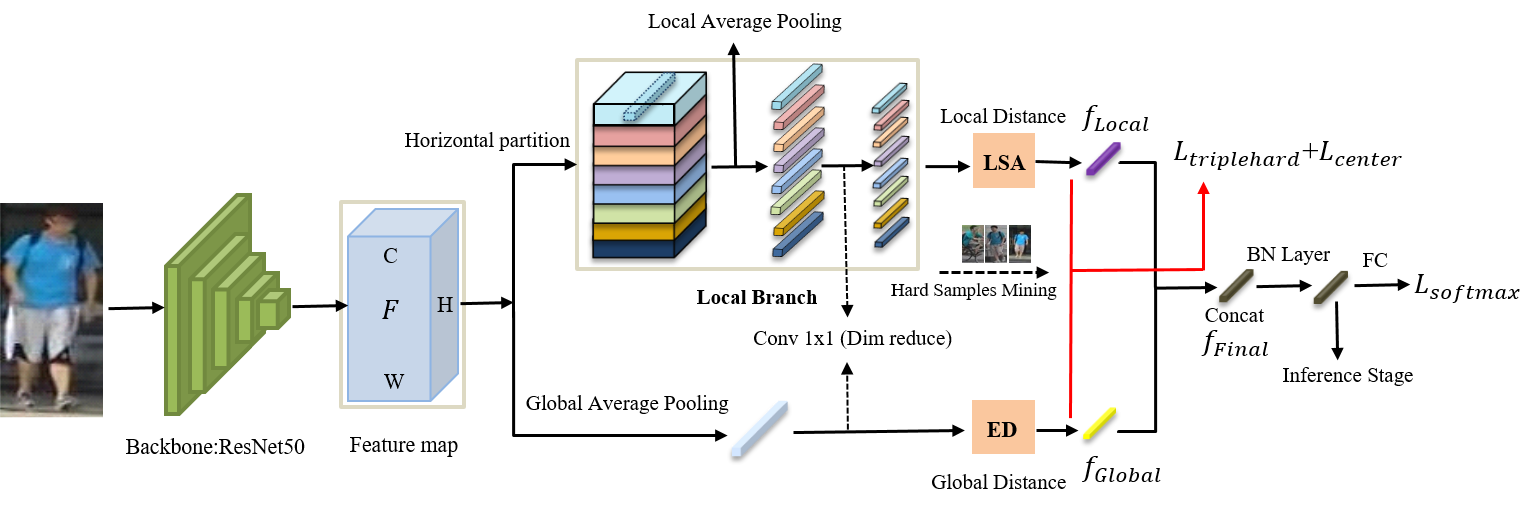}
	\end{minipage}	
	\caption{The network structure of GLDFA-Net. After the input image passes through the stacked convolutional layers of the Resnet50 backbone network, the global branch and the local branch share the feature map. For global features, we can directly perform global average pooling on the feature map and use Euclidean Distance (ED) to calculate the global distance. For local features, we first partition the feature map horizontally and obtain the feature vector of each horizontal strip through local average pooling. Then, we use LSA to calculate the local alignment distance and combine the global distance for triple loss with hard sample mining. Finally, we combine global features and local features as the final feature representation and pass through a fully connected (FC) layer and a softmax layer to achieve image classification.
	}
	\label{fig:3-1-Structure}
\end{figure*}
\subsection{Network Architecture}
In order to allow CNN to learn more discriminative features, we design a novel GLDFA-Net framework for person Re-ID tasks. \autoref{fig:3-1-Structure} shows the architecture of our proposed network. We use Resnet50, which has competitive performance and relatively simple architecture, as the backbone network to extract feature maps of pedestrians \cite{LUO201953AlignedReID,Sun2018BeyondPart,Bai2017Deep}. We first remove the average pooling layer and subsequent layers of Resnet50, and then divide the part after the ResBlock4 block into two independent branches, naming them Global Branch and Local Branch, respectively. For the global branch, we use Global Average Pooling (GMP) to convert the feature maps into a global feature vector. To reduce the dimension of local features to speed up the computational efficiency of distance, the local branch network and the global branch use Conv $1\times1$ to reduce the number of channels of the feature map. In this paper, the ResBlock4 block of the ResNet50 backbone network outputs a feature map with 2048-dim, and we use Conv $1\times1$ to reduce the number of channels of the feature map from 2048-dim to 256-dim. We change the step of the last spatial down-sampling of the Resnet50 backbone network from 2 to 1. When an image of size $384 \times 128$ is input, a feature map with a larger spatial size $(24 \times 8)$ can be obtained. \autoref{Table:table1} shows some details. For the local branch, we use horizontal average(max) pooling to evenly divide the output feature map into $k$ strips in the horizontal direction, and average all column vectors in the same strip into a single column vector $l_i$($i=1,2,3,...,k$, inspired by PCB \cite{Sun2018BeyondPart}, where the size of $k$ is set to 8), and the dimension of $l_i$ is set to 256. We learn local features independently on these strips.
\begin{table}[thp]\footnotesize
	\centering
	\caption{Comparison of component settings in GLDFA-Net. The size of the input image is set to $384 \times 128$. “Component” represents the name of the branch. “Map Size” refers to the size of the output feature of each branch. “Dimension” represents the dimension of the output feature map. “Description” represents the symbol of the output feature.} \label{Table:table1}
	\addtolength{\tabcolsep}{1.0pt}
	\begin{tabular*}{9.95cm}{cccc}
		\toprule
		Component & Map Size & Dimension &  Description  \\
		\midrule
		Backbone         &    12 x 4    & 2048  &         Null        \\
		Global-branch    &    24 x 8    & 256   &         ${f}_{Global}$      \\
		Local-branch     &    24 x 8    & 256 * 8 &       ${f}_{Local}^i\mid_{i=1}^8$   \\
		Final-branch     &    24 x 8    & 256 * (8+1) &   $f_{Final}=\left[f_{Local}\cdot f_{Global}\right]$ \\
		\bottomrule[0.75pt]
		\multicolumn{4}{p{6cm}}{\scriptsize}
	\end{tabular*}
\end{table}

In the testing phase, we reduce the global and local features to 256-dim and connected them to the final feature. In addition,the global branch and the local branch don't share weights, that is, their corresponding triple loss and classification loss are trained with independent weight constraints.

\begin{figure*}[t]
	\centering	
	\begin{minipage}[t]{0.95\linewidth}
		\centering
		\includegraphics[width=1.0\linewidth]{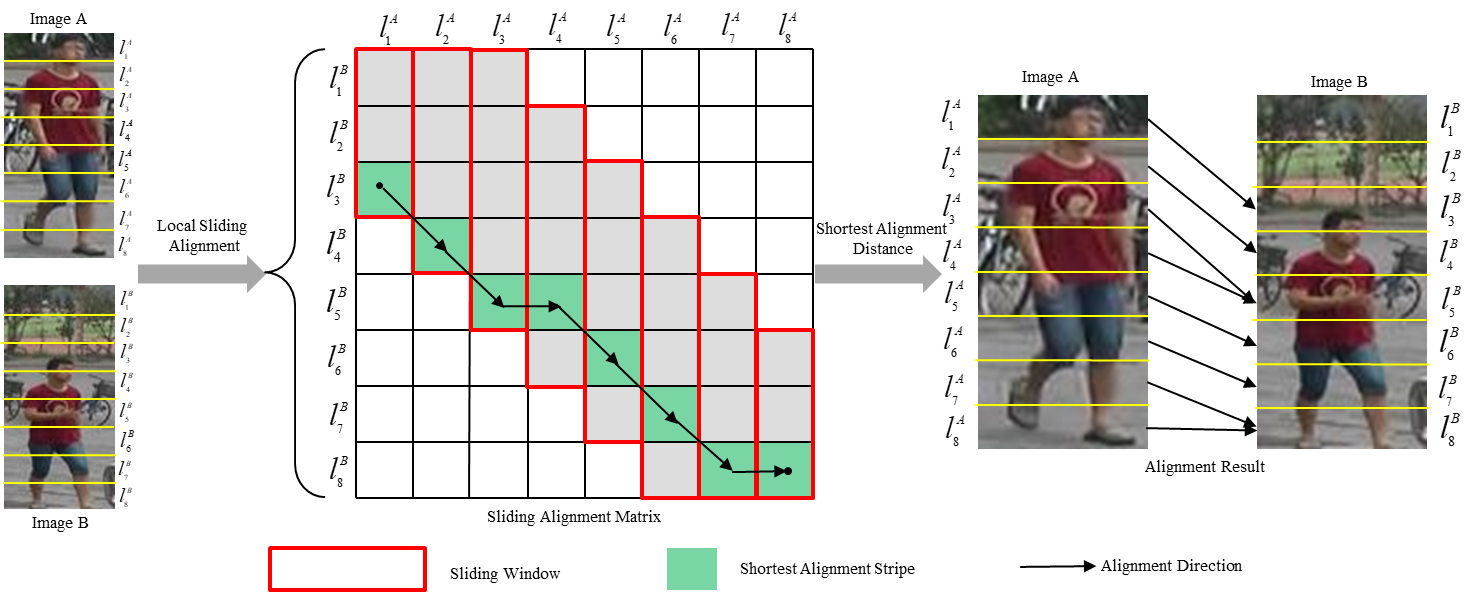}
	\end{minipage}	
	\caption{Schematic diagram of the Local Sliding Alignment (LSA). Firstly, we divide horizontal strips for pedestrians. Secondly, we set a sliding window for the partial strips from top to bottom. Finally, we calculate the shortest alignment distance from the horizontal strip in the window. The arrow direction represents the matching local feature, the solid arrow represents the shortest alignment distance, the dashed line represents the non-shortest alignment distance, and the alignment distance of the two images is the sum of the alignment distances of all local strips. 
	}
	\label{fig:3-1-LSA}
\end{figure*}

\subsection{Local Sliding Alignment}
In the training phase, we calculate the global distance of the global branch and the local distance of the local branch for the metric learning of the triple loss. In the inference stage, we use the local distance to calculate the similarity between the images. In this paper, we all use Euclidean distance to calculate the distance between the global branch and the local branch. We use ${f}_{G}^{A}$ and ${f}_{G}^{B}$ to represent the global features of images, respectively, and the Euclidean distance of the global feature can be expressed as:
\begin{equation}
{{G}_{dis}}= \left\|{f}_{G}^{A}-{f}_{G}^{B}\right\|_{2}
\end{equation}

We propose a dynamic alignment strategy called Local Sliding Alignment (LSA) to calculate local distance. The reason is that most traditional methods either use local hard alignment or introduce the key points of the pedestrian's body structure to assist the pedestrian's local alignment to calculate the local distance. However, when there are wrong boundary detection frames and occlusions of pedestrians, the local hard alignment method is less effective for mining hard samples, and the key point-based method needs to introduce additional computing resources.
\begin{algorithm} \small	
	\caption{Local Sliding Alignment} 
	\label{alg:LSA_Algorithm}
	\hspace*{0.02in} {\bf Input:}  	
	Images $A$ and $B$, sliding window size $W$\\ 	
	\hspace*{0.02in} {\bf Output:}  	
	Align distance $L^{dis}$
	\begin{algorithmic}[1]	
		\State{\textbf{Initialization:}${f}_{L}^{A}=\{{l}_{1}^{A},{l}_{2}^{A},{l}_{3}^{A},...,{l}_{k}^{A}\},{D}_{A}=\varnothing,{f}_{L}^{B}=\{{l}_{1}^{B},{l}_{2}^{B},{l}_{3}^{B},...,{l}_{k}^{B}\},$ \\ ${D}_{B}=\varnothing,{count}={0}$}
		\While{${True}$}
		\For{ $i \gets  1 ~\mathrm{to}~ k $ }
		\State{${d}_{align} \gets  {dis({l}_{i}^{A},{l}_{i}^{B})} $}
		\State{${w}_{up} \gets  max({1,i-W/2}) $}
		\State{${w}_{down} \gets  min({k,i+W/2})$}
		\For{ $j \gets  1 ~\mathrm{to}~ k $ } 
		\If{${j>={w}_{up}} \textbf{ and } {j<={w}_{dowm}}$} 		
		\State{${d}_{align} \gets  min({dis({l}_{i}^{A},{l}_{j}^{B})},{d}_{align}) $} 		
		\EndIf
		\EndFor 		
		\State{${D}_{A} \gets {d}_{align}$} 	
		\EndFor 
		\If{${count}>0$}
		\State{$L^{dis} \gets min(sum({D}_{A}),sum({D}_{B}))$}
		\State{break}
		\Else
		\State{${f}_{L}^{A} \leftrightarrow {f}_{L}^{B}$}
		\State{${D}_{B} \gets {D}_{A}$}
		\State{$count \gets count+1$}
		\EndIf
		\EndWhile
		
	\end{algorithmic}
\end{algorithm}	

LSA combines the ideas of sliding windows and dynamic programming. First, we divide the local branch feature map into horizontal stripes. Secondly, we set sliding windows for the horizontal stripes. Finally, we compute the local shortest alignment distance of the stripes in the sliding window to obtain the shortest alignment distance of the whole feature map. The details of the LSA implementation are presented in Algorithm  \ref{alg:LSA_Algorithm}. \autoref{fig:3-1-LSA} shows a schematic diagram of the LSA implementation process. 

Specifically, we use ${f}_{L}^{A}=\{{l}_{1}^{A},{l}_{2}^{A},{l}_{3}^{A},...,{l}_{k}^{A}\}$ and ${f}_{L}^{B}=\{{l}_{1}^{B},{l}_{2}^{B},{l}_{3}^{B},...,{l}_{k}^{B}\}$ to denote the local features of images ${A}$ and ${B}$ respectively, where $k$ represents the number of local stripes. In practice, we set the sliding window size to half of the number of stripes, i.e., ${W}=k/2$. The sliding step size ${S}$ is set to 1 by default. In addition, we perform a parameter sensitivity analysis on the number of local stripes $k$ and the sliding window size $W$ in \autoref{sec:ParametersAnalysis} to verify the validity of these parameter settings. The actual window size ${w}_{i}$ corresponding to the $i$-th stripe can be expressed as:
\begin{equation}\label{eqn:Lyapdisturb6} 
\begin{aligned}
{w}_{i}=\left [ max\left (1,i-W/2 \right ), min\left ( k,i+W/2 \right )    \right ] 
\end{aligned} 
\end{equation}

Where $i$ takes a range of values $1\le i\le k$, and ${k}$ represents the number of horizontal stripes, inspired by PCB \cite{Sun2018BeyondPart}, we set $k$ to $8$. The sliding window ${w}_{i}$ is the interval from $max\left (1,i-W/2 \right )$ to $ min\left ( k,i+W/2 \right )$. For example, the number of stripes $k=8$, then $W=4$, when $i=1$, the interval size of the window of image ${B}$ matched by the $i$-th stripe of image ${A}$ is ${w}_{i}=[1,3]$. When $i=4$, the interval size of the window of image ${B}$ matched by the $i$-th stripe of image ${A}$ is ${w}_{i}=[2,6]$. \autoref{fig:3-1-LSA} shows a schematic diagram of the LSA implementation process, where the red box indicates the sliding window ${w}_{i}$.

The set of sliding windows of image ${A}$ matching image ${B}$ can be represented as ${W}_{L}^{AB}=\{{w}_{1}^{A},{w}_{2}^{A},{w}_{3}^{A},...,{w}_{k}^{A}\}$.
The set of sliding windows of image ${B}$ matching image ${A}$ can be represented as
${W}_{L}^{BA}=\{{w}_{1}^{B},{w}_{2}^{B},{w}_{3}^{B},...,{w}_{k}^{B}\}$. We denote the shortest distance between the $i$-th stripe of image $A$ and the sliding window of image $B$ as ${d}_{i}^{AB}$, and the shortest distance between the $i$-th stripe of image $B$ and the sliding window of image $A$ as ${d}_{i}^{BA}$, then ${d}_{i}^{AB}$ and ${d}_{i}^{BA}$ can be expressed as:
\begin{equation}\label{eqn:Lyapdisturb5} 
\begin{aligned}
{{d}_{i}^{AB}}= min\left \{   \left \|  {l}_{i}^{A}-{l}_{i}^{B} \right \|_2 \mid   {l}_{i}^{A} \in{f}_{L}^{A}, {l}_{i}^{B} \in{w}_{i}^{B}   \right \} 
\end{aligned}
\end{equation}

\begin{equation}\label{eqn:Lyapdisturb9} 
\begin{aligned}
{{d}_{i}^{BA}}= min\left \{ \left\|{l}_{i}^{B}-{l}_{i}^{A}\right\|_{2} \mid {l}_{i}^{B} \in{f}_{L}^{B}, {l}_{i}^{A} \in{w}_{i}^{A}  \right \} 
\end{aligned} 
\end{equation}

We add the shortest alignment distance of each stripe of images ${A}$ and ${B}$ to the sets ${D}_{A}$ and ${D}_{B}$ respectively, ${D}_{A}=\left \{ {d}_{1}^{AB}, {d}_{2}^{AB},{d}_{3}^{AB},...,{d}_{k}^{AB} \right \} $ and ${D}_{B}=\left \{ {d}_{1}^{BA}, {d}_{2}^{BA},{d}_{3}^{BA},...,{d}_{k}^{BA} \right \} $. Then we get the shortest alignment distances ${L}^{dis}$ of images ${A}$ and ${B}$ are obtained , then ${L}^{dis}$ can be expressed as:  
\begin{equation}\label{eqn:Lyapdisturb10} 
\begin{aligned}
{L}_{dis}=min\left ( \sum_{i=1}^{k} {d}_{i}^{AB}, \sum_{i=1}^{k} {d}_{i}^{BA} \right) 
\end{aligned} 
\end{equation}

In the training phase, we use LSA to calculate the local alignment distance as the sample distance for triplethard loss. LSA can shorten the distance of samples with the same ID, effectively mining hard samples and guiding the model to learn more discriminative features. In the inference stage, we calculate the distance between two images as the sum of global and local distances for similarity measurement, which can further improve the accuracy of feature matching.

\begin{figure*}[t]
	\centering
	\subfigure[Bounding box detection error]{
		\includegraphics[width=0.45\linewidth]{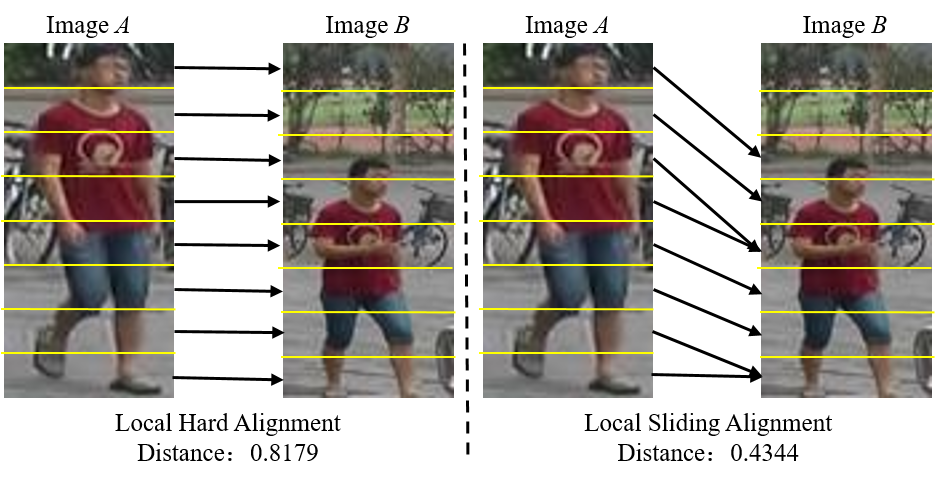}	
	}
	\centering
	\subfigure[Partial occlusion]{
		\includegraphics[width=0.45\linewidth]{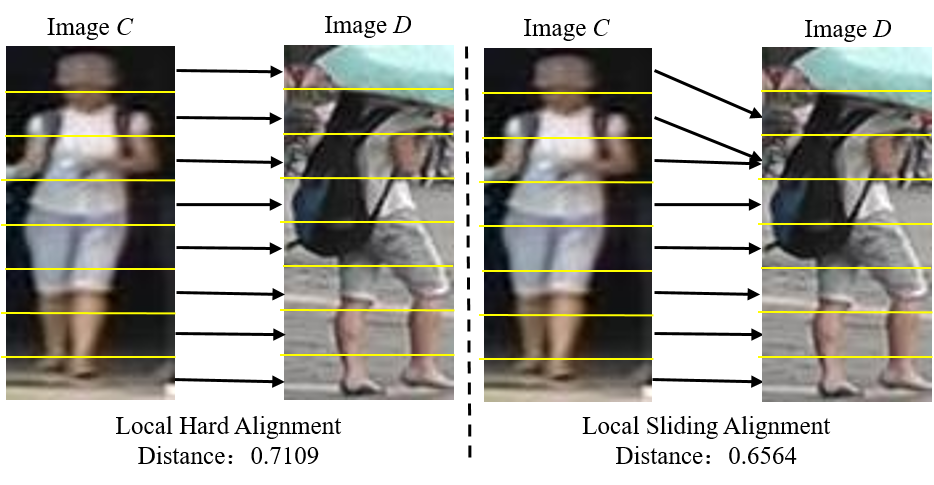}
	}
	\caption{Example of distance comparison between the local distance calculated by local hard alignment and local sliding alignment (LSA). In the case of bounding box detection errors and local occlusions, the local distances computed using LSA are much shorter than those with local hard alignment.}
	\label{fig:3-2-Align}
\end{figure*}

As shown in \autoref{fig:3-2-Align}, we get the local distance calculated by the local hard alignment method and the local sliding alignment (LSA) method respectively. In the case of bounding box detection errors and local occlusions, compared with the hard alignment strategy, the LSA we proposed can get a shorter alignment distance, which can make the distance among samples with the same ID shorter.
%
\subsection{Loss Functions}
To improve the model's ability to learn discriminative features, we joint classification loss and triplet loss to constrain the model. The classification loss usually connects a Fully Connected Layers (FC) for classification at the end of the network, and maps the feature vector of the picture to the probability space through the softmax activation function. The classification loss is also called ID loss. Therefore, the cross-entropy loss for multiple classifications of person Re-ID can be expressed as:
\begin{equation}
\mathcal{L} _{id}=\sum_{i=1}^{K}q(x_i)\log{p(y_i \mid x_i)} 
\end{equation}
where $K$ represents the number of ID categories of each batch of training samples, $q(x_i)$ represents the label of the sample image $x_i$, if $x_i$ is identified as $y_i$, then $q(x_i)=1$, otherwise $q(x_i)=0$. $p(y_i \mid x_i)$ is the probability that the image $x_i$ is predicted to be the category $y_i$ using the softmax activation function.

For the model to mine hard samples efficiently, We introduce an adaptive hard sample mining triplet loss (triplethard) proposed by Ristani et al. \cite{Ristani2018Multitarget}, which is an improved version of the original triplet loss. The triplehard loss function can be expressed as: 
\begin{equation} \label{eqn:Lyapdisturb3} 
\begin{aligned}
\mathcal{L} _{triplethard}=&[m+w_pd(x_a,x_p)-w_nd(x_a,x_n)]_+
\end{aligned} 
\end{equation}
\begin{equation}\label{eqn:Lyapdisturb2}
\begin{aligned}
w_p=\frac{\mathrm{exp} (d(x_a,x_p))}{\sum\limits_{x\in P(a)} \mathrm{exp} (d(x_a,x))} ,  
w_n=\frac{\mathrm{exp} (-d(x_a,x_n)}{\sum\limits_{x\in N(a)} \mathrm{exp} (d(x_a,x))}.
\end{aligned}
\end{equation}
where $\left [ \bullet  \right ] _{+}=max\left ( 0,\bullet  \right )$, $x_a$ is an anchor sample (Anchor), $x_p$ is a positive sample (Positive), $x_n$ is a negative sample (Negative), $x_a$ and $x_p$ have the same ID, $x_a$ and $x_n$ have different IDs, and $m$ is a hyper parameter set manually. By training the model, the distance between $x_a$ and $x_p$ in Euclidean space is closer than the distance between $x_a$ and $x_n$. We use the softmax function to adaptively assign weights $w_p$ and $w_n$ to positive and negative samples, respectively.

Although the triple loss can effectively improve the spatial distribution of features, it has a poor ability to constrain samples within the class. The center loss \cite{2016DiscriminativeWen} can minimize the distance of samples within a class and improve the compactness of samples of the same class. Therefore, we introduce a joint constrained model of the center loss and the triplethard loss to train. The center loss can be expressed as:
\begin{equation} 
\mathcal{L}_{center} =\frac{1}{2}\sum_{i=1}^{K}\left\|f_{t_i}-c_{y_i}\right\|_2^2
\end{equation}
where $k$ represents the number of ID categories of each batch of training samples, $y_i$ is the label of the batch training sample image $i$, and $c_{y_i}$ represents the class center of the deep feature $f_{t_i}$.

In our experiments, we calculate the triplethard loss and center loss of the global branch and local branch respectively. Therefore, the final triplethard loss and center loss can be expressed as:
\begin{equation} 
\mathcal{L}_{triplethard}^{'}=\mathcal{L}_{triplethard}^{{g}}+\mathcal{L}_{triplethard}^{{l} }
\end{equation}
\begin{equation} 
\mathcal{L}_{center}^{'}=\mathcal{L}_{center}^{{g}}+\mathcal{L}_{center}^{{l} }
\end{equation}

We connect the features $f_{Local}$ and $f_{Global}$ of the two branches as the final feature, which can be written as $f_{Final}=\left[f_{Local}\cdot f_{Global}\right]$, where $\left[\cdot\right] $ means concatenation. Finally, we use $f_{Final}$ to calculate the multi-class cross-entropy loss $\mathcal{L}_{id}$. Therefore, the final total loss $\mathcal{L}_{total}$ is a combination of the three losses and can be expressed as:
\begin{equation} 
\mathcal{L}_{total}=\mathcal{L}_{id}+\mathcal{L}_{triplethard}^{'}+\lambda \mathcal{L}_{center}^{'}
\end{equation}

Center loss simultaneously learns a center for deep features of each class and penalizes the distances between the deep features and their corresponding class centers, making up for the drawbacks of the triplet loss. The larger the weight of the center loss, the more effective the clustering of features will be. However, increasing the weight of center loss may make the retrieval performance of the model decrease. In our experiments, We follow the conclusion made in \cite{2016DiscriminativeWen,Luo2019Tricks}, in which the value of the weight $\lambda$ of the center loss has been studied and the authors conclude that $\lambda= 0.05$ gives better performance.
\section{Experimental Results}\label{ExperimentalResults}
\subsection{Implementation}
We adjust the size of all the images for training and testing to $384\times128$. Our method is implemented based on the open-source person Re-ID benchmark \cite{Luo2020NormalizationNeck}. We use weights that are pre-trained on ImageNet \cite{Deng2009ImageNet} to initialize the model. In the training phase, firstly, the pedestrian images are performed random horizontal flipping, random erasure and normalization to enhance the training data. Secondly, while training the network model, we use the triplethard loss and center loss better, we set the batch size to $32$, select $P$ samples with different identities randomly in each batch, and select $K$ images randomly for each identity from the training set. In our experiment, $P=8$ and $K=4$. The weights of the triplethard loss for global and local branches are both set to $0.3$. We choose Adam as the optimizer of the model. We set the size of epochs to $300$, where the learning rate is $3.5\times10^{-3}$ for the first $100$ epochs, the learning rate is $3.5\times10^{-4}$ between $100$ and $200$ epochs, and drops to $3.5\times10^{-5}$ after $200$ epochs, and the weight attenuation is set to $10^{-5}$. In addition, label smoothing (LS) \cite{Szegedy2016Rethinking} is used to improve the performance of the model. In the inference phase, we connect the feature vectors of the global branch and the local branch to generate the final feature representation. We use open-source re-ranking (RK) \cite{Zhong2017Reranking} technology to improve query results. Finally, our model is implemented on the PyTorch platform and NVIDIA 2080Ti GPU. Our experiments on all datasets share the same experimental settings as above.
\subsection{Datasets}
In this part, we will introduce the person Re-ID dataset used in this paper. Since Market1501 \cite{Zheng2015Scalable}, DukeMTMC-reID \cite{Zheng2017Unlabeled}, CUHK03 \cite{Wei2013HumanReidentification} and MSMT17 \citep{Wei2018PersonTransfer} are currently the mainstream datasets for person Re-ID tasks, we will evaluate the performance on these datasets.

\textbf{Market-1501.} Market-1501 is a large-scale person Re-ID data set released in 2015. This data set is collected by five high-resolution cameras and one low-resolution camera in front of the Tsinghua University supermarket. It contains a total of 32,668 images of 1,501 different pedestrians. The pedestrian detection box is marked by manual marking and automatic detector DPM. 

\textbf{DukeMTMC-reID.} DukeMTMC-reID is a subset of the multi-camera multi-target tracking data set DukeMTMC \cite{Ristani2016Performance}, which is also a person Re-ID data set. DukeMTMC-reID was collected by 8 static high-definition cameras on the campus of Duke University. It contains a total of 36,441 images of 1,812 different pedestrians. The size of each image is variable, and the pedestrian detection frame is manually labeled.

\textbf{CUHK03.} CUHK03 is a large-scale person Re-ID dataset. The images in CUHK03 are captured by ten cameras on the campus of the Chinese University of Hong Kong, including 1,360 different pedestrians for a total of 13,164 images. CUHK03 uses manual annotation and automatic detection to annotate pedestrian detection frames. The CUHK03 dataset is an improvement on the CUHK01\ cite{Li2013Locally} and CUHK02\cite{Li2014DeepReID} datasets by increasing the number of cameras and the captured images so that images from more cameras can be captured.

\textbf{MSMT17.}  MSMT17 is a large-scale person Re-ID dataset published in 2018 and is captured at the campus by fifteen cameras. The MSMT17 uses the pedestrian detector Faster R-CNN \citep{Ren2015FasterTowards} to automatically detect pedestrian-labeled frames. It contains 4,101 different pedestrian information with a total of 126,441 images, which is one of the large datasets of pedestrian and annotated images in the current person Re-ID task. The MSMT17 dataset can cover more scenes than earlier datasets and contains more views and significant lighting variations.

\textbf{Partial-Market and Partial-Duke.} The person images in the above dataset are carefully processed and contain almost the complete body of a person with limited detection of bounding box errors and occlusions of person. In this case,  the person Re-ID task can obtain high performance. To better verify the performance advantage of our method in processing part of the person Re-ID, we apply random erase and random crop operations to Market-1501 and DukeMTMC-reID, where some images are randomly erased 10\%-30\% in the vertical direction for modeling occlusion and some images are randomly cropped 20\%-30\% in the vertical direction for simulate detection of bounding box errors. We obtain the Partial-Market and Partial-Duke of partial person Re-ID datasets. We also do an ablation study on Partial-Market and Partial-Duke to verify the effectiveness of our method in solving partial person Re-ID.
\subsection{Ablation Study}
To further study the effectiveness of global-local feature combination and LSA strategy in GLDFA-Net, we design several ablation experiments with different settings on the Market-1501 dataset. The remaining parameters for each comparison experiment are the same as GLDFA-Net in Section 4.1. In addition, we use Resnet50 as a benchmark for comparing global and local feature networks, use Softmax as the classification loss function by default, and ensure that the selected model can get the best experimental performance under specified conditions. \autoref{Table:table_ablation} shows the results of ablation experiments related to GLDFA-Net components but with different settings, from which we can get the following conclusions:

\textbf{The impact of multiple networks:} Usually, the global branch network learns coarse-grained global information about the pedestrian. The local branching network learns detailed information about the pedestrian part. We believe that combining global and local branches can achieve better performance. Therefore, we use Resnet50 as the benchmark network to train global branch, local branch and global-local networks separately. We use Softmax classification loss and Triplethard loss training models to explore the impact of different network structures on model performance. We can observe from \autoref{Table:table_ablation} that the experimental results of the local branch network (Local + TH) are slightly better than the global branch network (Global + TH), and the global-local (Global-Local + TH) combined network is significantly better than the previous two. GLDFA-Net can achieve 92.6\% Rank-1 accuracy and 84.2\% mAP without using LSA (w/o LSA). Compared with the independent network, the global-local network can learn more discriminative features. Global and local branches interact with each other during the training process to learn complementary global and local features as discriminative information for pedestrian identity.

\textbf{The impact of local feature alignment:} There are often challenges such as pedestrian bounding box errors and partial occlusion, which will limit the alignment of local features and affect the accuracy of person Re-ID. To verify the effectiveness of the Local Sliding Alignment (LSA) strategy for partial Re-ID, we conduct a comparative experiment on the Market1501 dataset. L+LSA refers to the combined training model using local branch and LSA. GL+LSA refers to the combined training model using global-local and LSA, and our loss function is set to a combination of ID loss and triplethard loss. We compared the experimental results of L and GL in \autoref{Table:table_ablation}. We found the accuracy of Rank-1 using LSA increased by 2.2\% and 2.4\%, and mAP increased by 1.8\% and 1.1\%, respectively.  Our method GLDFA-Net can achieve the experimental performance of Rank-1/mAP=86.1\%/94.8\% when using LSA. Compared with local and global, LSA can greatly improve performance. LSA not only helps to mine samples in the training phase but also helps to calculate the alignment distance between samples in the inference phase. In general, the LSA we proposed is very effective for people Re-ID tasks, and it is of great significance for studying large-scale person Re-ID tasks in the real world.
\begin{table}[thp]\footnotesize
	\centering
	\caption{Ablation experiments on the Market1501 dataset. 'TH' refers to the triplethard loss. 'G' refers to the use of only a global feature learning network and triplethard loss. 'L' means that only the local feature learning combined network and triplethard loss. 'GL' refers to the use of a global-local feature learning network and triplethard loss. 'LSA' refers to the use of a local sliding alignment strategy to match the local features.} \label{Table:table_ablation}
	\addtolength{\tabcolsep}{1.0pt}
	\begin{tabular*}{7.95cm}{ccccc}
		\toprule
		Methods & mAP  &  R-1 & R-5 & R-10\\
		\midrule
		ResNet50             & 70.6  &  83.7 & 92.8 & 96.3  \\
		ResNet101            & 73.4  &  85.3 & 93.0 & 97.2  \\
		ResNet50+TH          & 73.8  &  86.8 & 93.2 & 97.4  \\
		\midrule
		G(Global+TH)         & 77.6  &	89.7 & 93.4 & 97.5  \\
		L(Local+TH)          & 78.9  &	90.1 & 94.0 & 97.9  \\
		GL(Global-Local+TH)  & 80.5  &	90.4 & 94.6 & 98.0  \\
		L+LSA                & 80.7  &	92.3 & 95.3 & 98.2  \\
		GL+LSA               & 81.6	 &  92.8 & 95.6 & 98.2  \\
		\midrule
		GLDFA-Net w/o TH     & 82.7  &	91.9 & 94.3 & 97.4  \\
		GLDFA-Net w/o LSA    & 84.2  &	92.6 & 95.0 & 97.8  \\
		GLDFA-Net            & \textbf{86.1}  &	\textbf{94.8}& \textbf{97.2} &\textbf{98.4}  \\		
		\bottomrule[0.75pt]
	\end{tabular*}
\end{table}

\textbf{The impact of triplethard loss:} In the person Re-ID task, we use ID loss for representation learning and triplethard loss for metric learning. To verify the enhancement effect of the classification loss and Triplethard loss joint training on the model performance, we use the Resnet50 benchmark network and GLDFA-Net to conduct experiments respectively. It is observed from the experimental results in \autoref{Table:table_ablation}, we can see that the Rank-1/mAP performance of the ResNet-50 baseline model improves by 3.1\%/2.6\%, and the Rank-1/mAP performance of GLDFA-Net is improved by 3.9\%/3.4\%. Compared to the benchmark model, triplethard loss improves the performance of the GLDFA-Net model even more. The triplethard loss improves the performance of the GLDFA-Net model more compared to the benchmark model. Compared to using the Softmax classification loss only, triplethard loss can learn the mapping relationship from the original image to the embedding space and can effectively improve the spatial distribution of features.
\begin{figure*}[t]
	\centering
	\subfigure[Random erasing]{
		\includegraphics[width=0.45\linewidth]{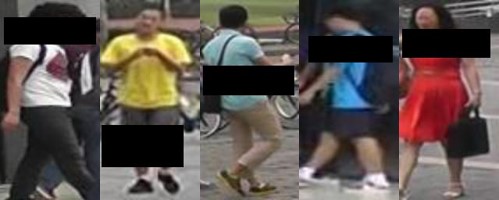}	
	}
	\centering
	\subfigure[Random cropping]{
		\includegraphics[width=0.45\linewidth]{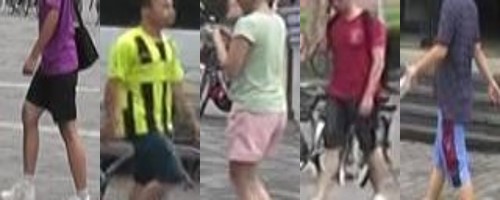}
	}
	\caption{Sampled person images of Partial-Market}
	\label{fig:4-4-Erasing-Crop}
\end{figure*}
\begin{table}[thp]\footnotesize
	\centering
	\caption{Comparison experiments on Partial-Market and Partial-Duke datasets.}
	\label{Table:table_ablation_partial}
	\addtolength{\tabcolsep}{-2.0pt}
	\begin{tabular*}{7.45cm}{lccccc}
		\toprule
		\multirow{2}{*}{Methods} & \multicolumn{2}{c}{Partial-Market}  & & \multicolumn{2}{c}{Partial-Duke}   \\
		\cmidrule[0.5pt]{2-3}\cmidrule[0.5pt]{5-6}   & mAP & Rank-1 & & mAP & Rank-1       \\
		\midrule
		ResNet50                                  & 42.8  & 60.5   &  &   39.7  &   52.3 \\
		
		AWTL\cite{Ristani2018Multitarget}         & 43.6  & 61.4   &  &   40.9  &   53.2 \\
		
		PCB\cite{Sun2018BeyondPart}               & 46.4  & 62.1   &  &   41.8  &   53.9 \\
		PCB+RPP\cite{Sun2018BeyondPart}           & 50.1  & 67.4   &  &   43.3  &   55.4 \\
		
		SCPNet\cite{2018SCPNetFan}                & 52.7  & 68.0   &  &   44.8  &   56.2 \\
		
		AlignedReID++\cite{LUO201953AlignedReID}  & 52.9  & 68.6   &  &   46.1  &   57.0 \\
		
		Pyramid\cite{Fu2018Horizontal}            & 53.2  & 68.9   &  &   46.6  &   58.4 \\ 
		
		MGN\cite{Wang2018LearningDiscriminative}  & 53.5  & 69.1   &  &   47.0  &   58.6 \\
		\midrule 
		\multicolumn{1}{c}{\textbf{GLDFA-Net}}   & \textbf{54.6} & \textbf{72.4} & &  \textbf{48.2} & \textbf{60.9} \\
		\bottomrule[0.75pt]
	\end{tabular*}
\end{table}

\textbf{Evaluation on partial person Re-ID:} We argue that the person images in Market-1501 and DukeMTMC-ReID are manually processed and contain almost the complete body of a person, with limited detection of bounding box errors and occlusions. To better evaluate the effectiveness of GLDFA-Net, we compare the experimental performance of our method with other classical methods on Partial-Market and Partial-Duke. \autoref{fig:4-4-Erasing-Crop} shows some example images, which are partial person Re-ID images of Partial-Market.

To verify the effectiveness of GLDFA-Net in partial person Re-ID, we conducted Comparison experiments on Partial- Market and Partial- Duke, and the evaluation results are shown in \autoref{Table:table_ablation_partial}. As a whole, the rank-1 accuracy and mAP of both baseline (Resnet50) and GLDFA-Net on Partial- Market and Partial- Duke decreased. The rank-1 accuracy of baseline on Market1501 was 83.7\%, but its performance on Partial-Market decreased to 60.5\%. The rank-1 accuracy of GLDFA-Net is 94.8\% and 90.1\% for Market-1501 and DukeMTMC-reID, respectively, but its performance drops to 72.4\% and 60.9\% for Partial-Market and Partial-Duke, respectively. In particular, our proposed method GLDFA-Net can achieve better performance on Partial-Market and Partial-Duke datasets compared to PCB \cite{Sun2018BeyondPart}, SCPNet \cite{2018SCPNetFan}, AlignedReID++ \cite{LUO201953AlignedReID}, Pyramid \cite{Fu2018Horizontal}  and MGN \cite{Wang2018LearningDiscriminative}, which is since GLDFA-Net employs a local dynamic feature alignment strategy to effectively suppress spatial misalignment. In overview, our proposed GLDFA-Net is more effective for partial person Re-ID tasks, which is more suitable for large-scale person Re-ID in bounding box errors and partial occlusion scenarios.
\subsection{Parameters Analysis}\label{sec:ParametersAnalysis}
\begin{figure*}[t]
	\centering
	\subfigure[Impact of the number of stripes $k$]{
		\includegraphics[width=0.45\linewidth,height=0.4\linewidth]{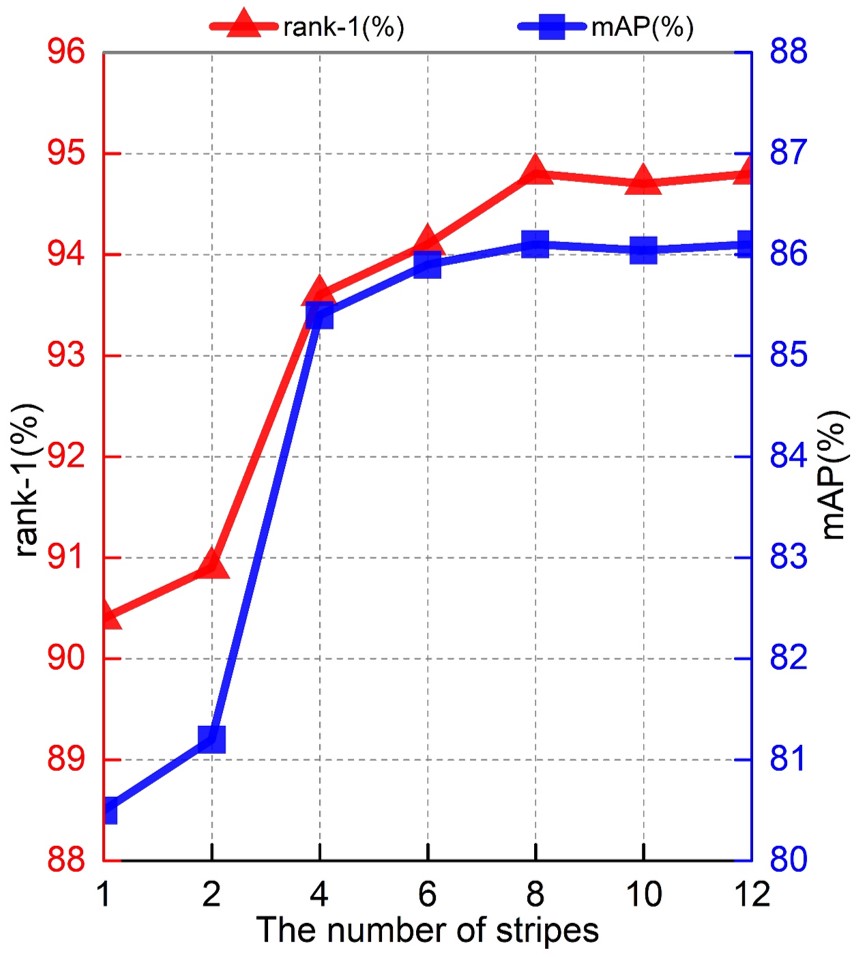}
	} 
	\quad
	\centering
	\subfigure[Impact of the size of sliding windows $W$]{
		\includegraphics[width=0.45\linewidth,height=0.4\linewidth]{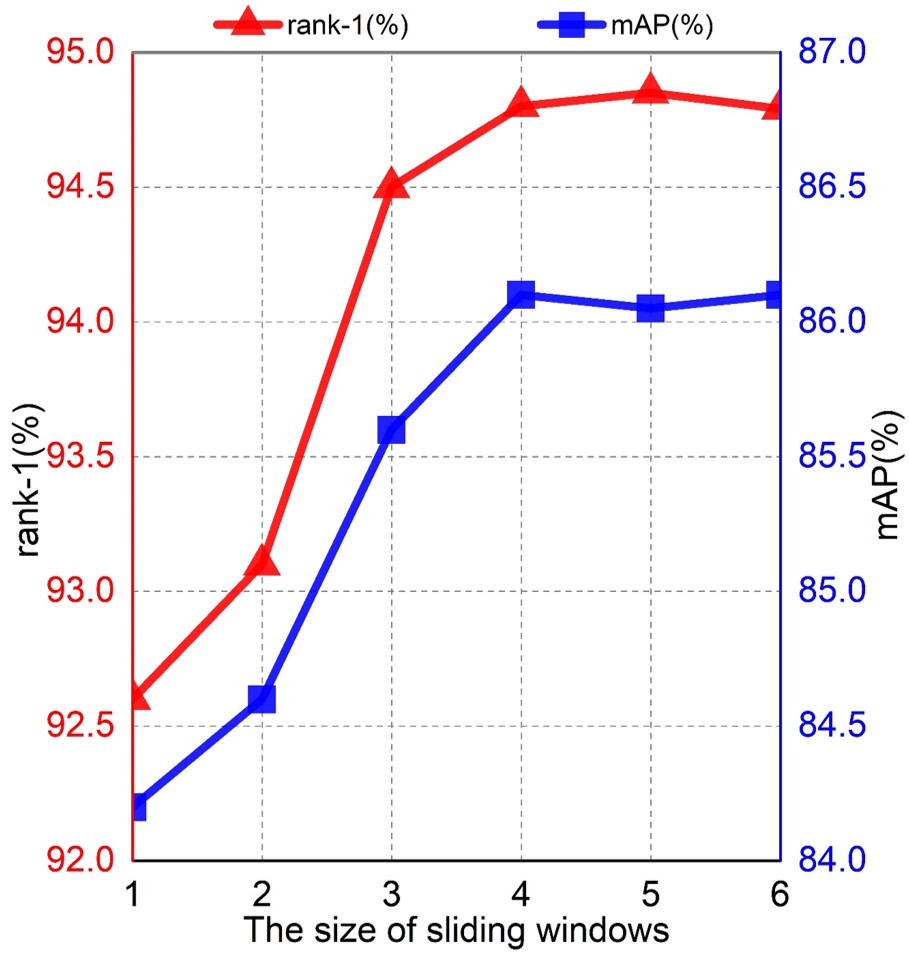}
	}
	\caption{Parameter analysis. (a):The impact of the number of stripes $k$. We set the size of the sliding window to be half the number of stripes. (b):The impact of the size of sliding windows $W$. We set the number of stripes is $8$.}
	\label{fig:4-4-Stripes-SW}
\end{figure*}
\begin{figure*}[t]
	\centering
	\subfigure[Alignment results for the number of stripes $k=8$]{
		\includegraphics[width=0.45\linewidth]{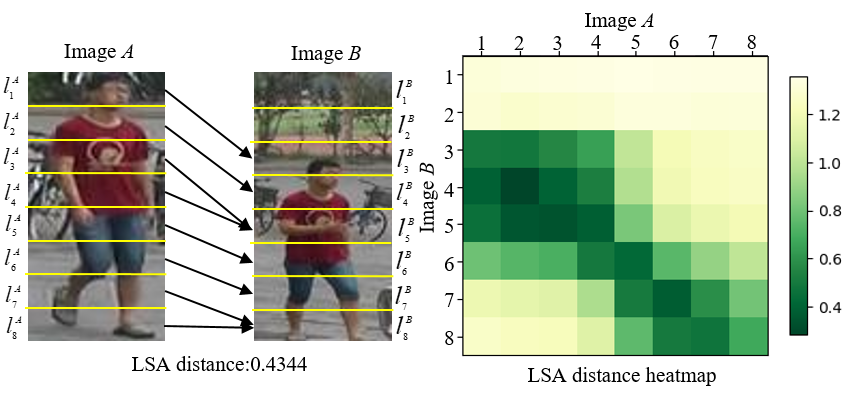}
	} 
	\centering
	\subfigure[Alignment results for the number of stripes $k=6$]{
		\includegraphics[width=0.45\linewidth]{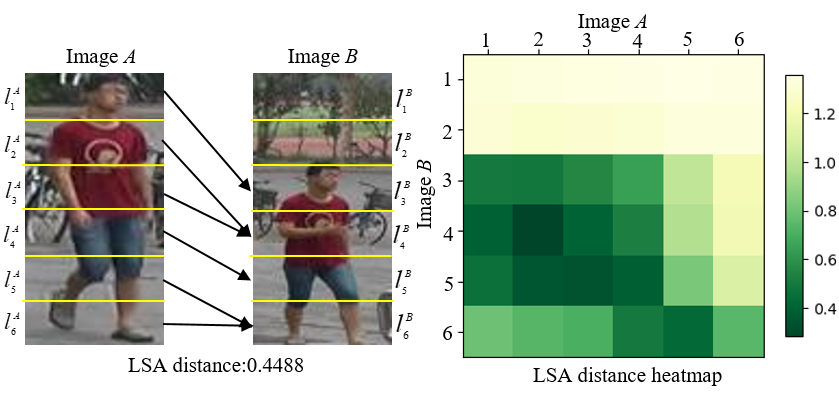}
	}
	\caption{Example alignment results for different number of stripes. (a): Alignment results for the number of stripes $k=8$. (b) :Alignment results for the number of stripes $k=6$. We also show the distance maps/matrices of the local features of the two images.}
	\label{fig:4-4-Stripes-Numbers}
\end{figure*}
We analyze some important parameters of GLDFA-Net (and with LSA) introduced in Section 3.2 on Market-1501. Once optimized, the same parameters are used for all datasets in the train and inference stage.

\textbf{The impact of the number of stripes $k$:} The number of stripes $k$ determines the granularity of the local features. When $k=1$, the features learned by the neural network are global, and we usually assume that the retrieval accuracy of Re-ID increases as $k$ increases. However, the retrieval accuracy does not necessarily increase all the time. We set the size of the sliding window to half of the number of stripes. As shown in \autoref{fig:4-4-Stripes-SW}(a), the retrieval accuracy rank-1 and mAP increase with the number of stripes before $k=8$. The alignment results for the different number of stripes are shown in \autoref{fig:4-4-Stripes-Numbers}, where the alignment distance for $k=$8 is smaller than that for $k=6$.  After $k=8$, the retrieval accuracy rank-1 and mAP hardly fluctuate, probably because the sliding window $W$ increases while the stripe number $k$ increases, and the local feature alignment distance does not change much. In real-world applications, the increase in the number of stripes affects the execution efficiency of the model, and we suggest that the number of stripes $k=8$.

\textbf{The impact of the size of sliding windows $W$:} In \autoref{fig:4-4-Stripes-SW}(b), we set the number of stripes $k=8$ and explore the effect of the size of the sliding window $W$ on the retrieval accuracy. When $W=1$, it is the traditional way of hard alignment of stripes. Before $W=4$, the retrieval accuracy rank-1 and mAP increase with the increase of sliding window $W$. The reason is that the stripes of one image can match more stripes within the sliding window of another image as $W$ increases, and getting a shorter alignment distance to improve the retrieval accuracy. After $W=4$, the retrieval accuracy rank-1 and mAP hardly change, the reason may be that the shortest alignment distance within the window remains the same. In practical applications, we propose that $W=k/2$ can align almost most of the body parts.

\begin{figure*}[t]
	\centering
	\begin{minipage}[t]{0.85\linewidth}
		\centering
		\includegraphics[width=1.0\linewidth]{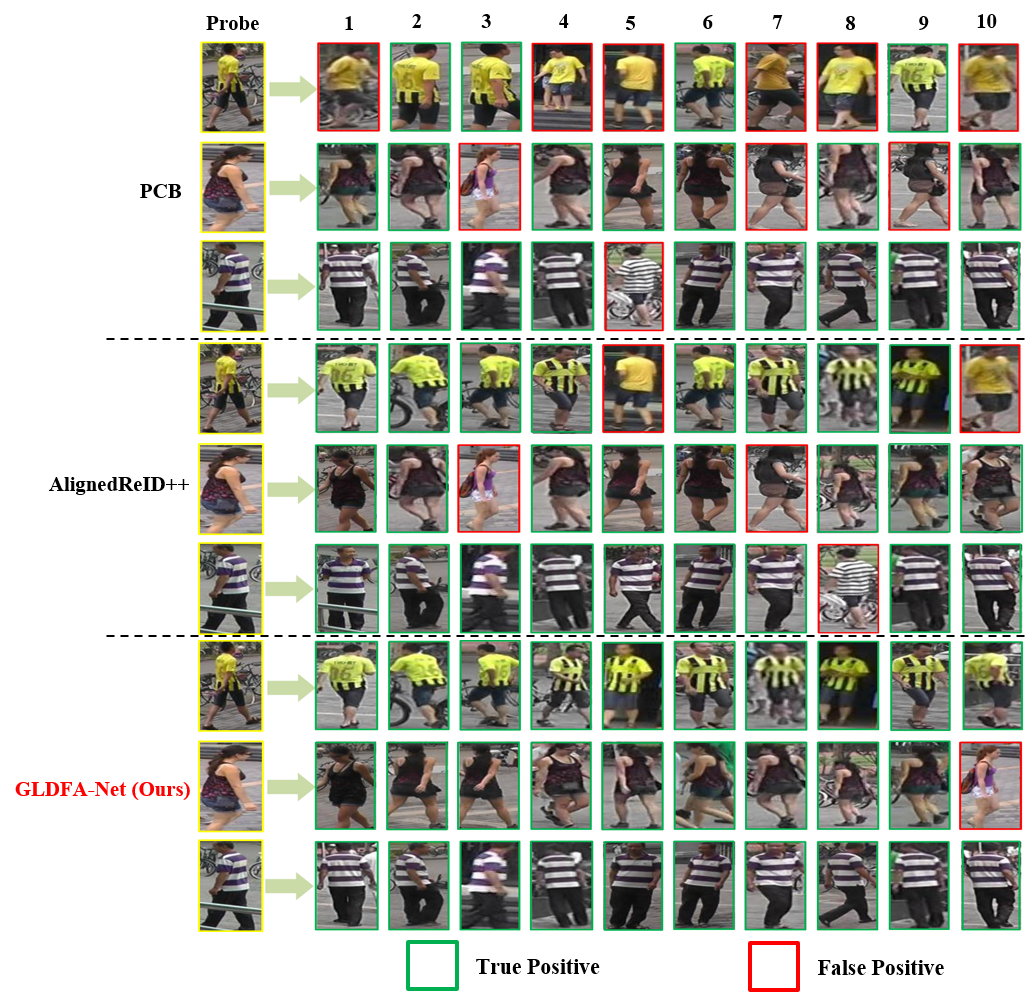}
	\end{minipage}
	\caption{Qualitative comparisons of three randomly sampled query results on the Market-1501 dataset. From top to bottom are the query results of PCB\cite{Sun2018BeyondPart}, AlignedReID++\cite{LUO201953AlignedReID} and our proposed method. The left column is the query image. The images on the right side are the top-10 ranking retrieved results. The images with green borders belong to the same identity as the given query image, while the one with a red border shows the incorrect identity. 
	}
	\label{fig:4-3-QueryResults}
\end{figure*}
\subsection{Comparison with State-of-the-Art Methods}
We compared GLDFA-Net with recent methods on the three data sets of Market1501, DukeMTMC-ReID and CUHK03 to prove that GLDFA-Net is superior to other existing methods. To better illustrate the comparison results on each data set, we will introduce the following in detail:

\textbf{Market1501:} \autoref{Table:table4} shows the person Re-ID results on Market1501. We divide these methods into two groups: global-based methods and local-based methods. From the overall experimental results, the performance of local-based methods is generally higher than that of methods with global features, such as SVDNet \cite{Sun2017SVDNet}, Mancs \cite{Wang2018Mancs}, Triplet Loss \cite{Hermans2017InDefense}. For the local-based methods, in the single query mode, PCB+RPP \cite{Sun2018BeyondPart} only uses the local horizontal stripe partition and Refined Part Pooling(RPP) to obtain better results, but this type of method often ignores the impact of global features on discriminative feature mining. We combine coarse-grained global features and fine-grained local features. Our GLDFA-Net achieves Rank-1/mAP=94.8\%/86.1\%, which improves the accuracy of Rank-1 by 1.0\% and 4.5\% on mAP compared to PCB+RPP. After using re-ranking, our experimental results can reach Rank-1/mAP=95.6\%/93.5\%, which is much better than other current methods. On one hand, we use a combination of global and local features to enrich the discriminative features of pedestrians, which helps to improve the accuracy of the model. On the other hand, although RPP obtains local discriminative features by refined pooling of local stripes, it still belongs to the category of hard alignment, which is less effective than GLDFA-Net in mining hard samples in the case of bounding box detection error or partial occlusion.
\begin{table*}[thp]\footnotesize
	\centering
	\caption{Comparison with the state-of-the-art methods in terms of R-1(Rank-1) and mAP on the Market-1501, Duke-reID(DukeMTMC-reID) and CUHK03 datasets. Group 1: the methods only use global features. Group 2: the methods using local features. The bold font represents the best result. RK stands for the method of using re-ranking \cite{Zhong2017Reranking}.}
	\label{Table:table4}
	\addtolength{\tabcolsep}{0.6pt}
	\begin{tabular*}{11.8cm}{lccccccccccccccccccccccccccccccc}
		\toprule 
		\multirow{3}[2]{*}{Methods} & \multicolumn{2}{c}{\multirow{2}[1]{*}{Market-1501}} & \multicolumn{2}{c}{\multirow{2}[1]{*}{DukeMTMC-reID}} & \multicolumn{4}{c}{CUHK03} \\
		\cmidrule{6-9}    \multicolumn{1}{c}{} & \multicolumn{2}{c}{} & \multicolumn{2}{c}{} & \multicolumn{2}{c}{Detected} & \multicolumn{2}{c}{Labeled} \\
		\cmidrule{2-9}    \multicolumn{1}{c}{} & \multicolumn{1}{c}{mAP} & \multicolumn{1}{c}{R-1} & \multicolumn{1}{c}{mAP} & \multicolumn{1}{c}{R-1} & \multicolumn{1}{c}{mAP} & \multicolumn{1}{c}{R-1} & \multicolumn{1}{c}{mAP} & \multicolumn{1}{c}{R-1} \\

		\midrule
		Triplet Loss\cite{Hermans2017InDefense}     &    69.1  & 84.9  &   58.8  &  76.7        &   -     &   -      & -   & -     \\
		SVDNet\cite{Sun2017SVDNet}                  &    62.1  & 82.3  &    71.8  &   84.9 &          37.3  &   41.5      & 37.8   & 40.9   \\
		$\ast$ Mancs\cite{Wang2018Mancs}            &    82.3  & 93.1   &   76.4  &   86.4 &         60.5   & 65.5        & 63.9   &  69.0    \\
		\midrule
		$\Phi $ PDC\cite{Su2017PoseDriven}             &    63.4  & 84.1  &     -  &   - &        -   & -       & -   & -     \\
		$\Phi $ GLAD\cite{Wei2019GLAD}                 &      73.9  & 89.9  &    -  &   - &      -   & -   &     -   & -     \\
		$\ast$ HA-CNN\cite{Li2018Harmonious}        &    75.5  & 91.2  &   63.8  &   80.5 &       38.6   & 41.7   &    41.0   & 44.4 \\
		$\Phi $ Pose-transfer\cite{Liu2018PoseTransferrable}  &    56.9  & 78.5  &   48.1  &   68.6       &  38.7   & 71.6      & 42.0 & - \\
		$\dagger$ PCB\cite{Sun2018BeyondPart}       &    77.4  & 92.3  &   66.1  &   81.8 &       54.2   & 61.3   &    -   & -   \\
		$\dagger$ PCB+RPP\cite{Sun2018BeyondPart}   &    81.6  & 93.8  &   69.2  &   83.3 &        57.5   & 63.7     & -   & -  \\
		$\dagger$ HPM\cite{Fu2018Horizontal}        &    82.7  & 94.2  &   74.3  &   86.6      &  57.5   & 63.9    & -   & -   \\
		$\dagger$ MGN\cite{Wang2018LearningDiscriminative}        &      -    &   -   &    -    &     -  &       66.0   & 68.0   &    67.4   & 68.0   \\
		$\ast$ MHN(IDE)\cite{Chen2019Mixed}         &    83.6  & 93.6  &    75.2  &   87.5 &       61.2   & 67.0   &   65.1   & 69.7   \\
		$\dagger \ast$ VPM\cite{Sun2019Perceive}    &    80.8  & 93.0  &   72.6  &   83.6 &        -     &  -     &      -    &   -    \\
		$\ast$ PL-Net\cite{Yao2019DeepRepresentation}   &    69.3  & 88.2  &     -  &      - &      -     &  -     &   -     &   -  \\
		$\dagger \ast$ AANet\cite{Tay2019AANet}	        &    83.4  & 93.9  &  74.3  & \textbf{87.6} &    -     &  -     &     -     &   -     \\
		$\dagger$ AlignedReID++\cite{LUO201953AlignedReID}  &      79.1  & 91.8  &   69.7  &   82.1 &      59.6   & 61.5   &    -     &   -  \\
		$\ast$ DHA\cite{2020DHAWang}                &     76.0  & 91.3  &    64.1  &   84.3 &         -     &  -     &     -    &   -           \\
		$\Phi $ PPA+TS\cite{2021PersonLi}              &    79.6  & 92.4  & 71.8  &   85.1 &    62.0   & 64.6   &  66.3   & 69.2  \\
		$\Phi $ MBRAN\cite{2020MultiFang}              &   83.8  & 94.6  &   76.9  &   87.1 &      -     &  -     &     -     &   -   \\
		$\Phi \ast$ FGSAM \cite{2020FineZhou}          &    85.4  & 91.6  &   71.4  &   85.9 &       -     &  -     &     -     &   -    \\
		\midrule
		\multicolumn{1}{c}{\textbf{GLDFA-Net}}          & \textbf{86.1 } & \textbf{94.8} & \textbf{76.9} & 87.5     &  \textbf{68.4}   &   \textbf{72.3}  &  \textbf{68.9}   &   \textbf{73.1}     \\
		\multicolumn{1}{c}{\textbf{GLDFA-Net(RK)}}      & \textbf{93.5}  & \textbf{95.6} & \textbf{89.5 } & \textbf{90.1}    & \textbf{70.3} &    \textbf{73.6}    & \textbf{72.1}  &   \textbf{76.5}    \\
		\bottomrule[0.75pt]
		\multicolumn{9}{c}{\scriptsize  $\dagger$ Stripes related, $\ast$ Attention related, $\Phi $ Pose or human parsing related.}
	\end{tabular*}
\end{table*}

\autoref{fig:4-3-QueryResults} shows the top-10 ranking retrieved results of three methods on the Market-1501 dataset.  From top to bottom are the query results of PCB\cite{Sun2018BeyondPart}, AlignedReID++\cite{LUO201953AlignedReID} and our proposed method. These retrieved images are all from the Gallery collection and captured by different cameras. From the retrieval results, the GLDFA-Net method is significantly better than the first two. The first class of pedestrians has a large variety of posture or gait, our method can still get all the correct retrieval results by aligning the invariant features of the body parts. Errors in the pedestrian boundary detection box in the second class lead to easy loss of important information or introduce background information, which our method can effectively mitigate by dynamically aligning the pedestrian body parts. The third class shows the retrieval results for partially occluded pedestrians. Although the occluded area not only loses a lot of important information but also introduces additional noise, our alignment method is still able to obtain high retrieval results. The retrieval results show strong robustness, except for the last matching error. We attribute this surprising result to the effect of the dynamic alignment of local features that reflect the robustness of their identity.

\textbf{DukeMTMC-ReID:} As can be seen from \autoref{Table:table4}, GLDFA-Net also shows better performance on the more challenging DukeMTMC-reID dataset. Without re-ranking, our method can achieve an accuracy of Rank-1/mAP =87.5\% /76.9\%. Compared to AlignedReID++ \cite{LUO201953AlignedReID}, our method improves Rank-1 by 5.4\% and mAP by 7.2\%. Although AlignedReID++ can align local stripes using the full alignment strategy, when there is an error in the detection of the bounding box, the wrong stripe matching result is also calculated into the alignment distance and affects the final similarity. LSA introduces the idea of sliding window and dynamic programming, which effectively mines hard samples by adaptively aligning local features and metric similarity.

\textbf{CUHK03:} As shown in \autoref{Table:table4}, GLDFA-Net reached Rank-1/mAP = 73.1\%/68.9\% on the manually labeled CUHK03 dataset. In particular, Rank-1/mAP = 72.3\%/68.4\% is achieved on the automatically detected CUHK03 dataset, which has exceeded the experimental results of most published methods. The automatically detected CUHK03 dataset is prone to incorrect pedestrian bounding boxes, which will produce missing or redundant information of body parts. GLDFA-Net uses the dynamic alignment of the local features, which can effectively avoid the influence of these factors and can produce advanced accuracy. 

\begin{table}[htbp]
  \centering
  \caption{Comparison with the state-of-the-art methods in terms of R-1(Rank-1), R-1(Rank-5) and mAP on the MSMT17 datasets.}
    \begin{tabular}{p{12.125em}ccp{4.19em}}
    \toprule
    \multirow{2}[4]{*}{{Methods}} & \multicolumn{3}{c}{{MSMT17}} \\
\cmidrule{2-4}    \multicolumn{1}{c}{} & \multicolumn{1}{p{4.19em}}{{mAP}} & \multicolumn{1}{p{4.19em}}{{Rank-1}} & {Rank-5} \\
    \midrule
    PDC\cite{Su2017PoseDriven}            & 29.7  & 58.0  & 73.6    \\
    GLAD\cite{Wei2019GLAD}                & 34.0  & 61.4  & 76.8    \\
    OSNet\cite{Zhou2019OmniScale}         & 52.9  & 78.7  & -      \\
    Auto-ReID\cite{2019Auto-reidZhou}     & 52.5  & 78.2  & 88.2   \\
    DG-Net\cite{2019DG-NetZheng}        & 52.3  & 77.2  & 87.4   \\
    BAT-Net\cite{2019BAT-NetFang}       & 56.8  & 79.5  & 89.1   \\
    SFT\cite{2019SFTLuo}               & 58.3  & 79.0  & 85.5    \\
    SAN\cite{2020SemanticsJin}           & 55.7  & 79.2  & -      \\
    \midrule
    \textbf{GLDFA-Net} & \textbf{58.7} & \textbf{79.8} & \textbf{89.1} \\
    \textbf{GLDFA-Net(RK)} & \textbf{69.4} & \textbf{83.0} & \textbf{91.8} \\
    \bottomrule
    \end{tabular}%
  \label{Table:Table_MSMT17}%
\end{table}%

\textbf{MSMT17:} The MSMT17 dataset is a large-scale person Re-ID dataset published in 2018 and is captured at the campus by fifteen cameras. It can cover more scenes than earlier datasets and contains more views and significant lighting variations. \autoref{Table:Table_MSMT17} shows the comparison of the proposed GLDFA-Net with the state-of-the-art methods on the MSMT17 dataset. We proposed GLDFA-Net reached Rank-1/mAP = 79.8\%/58.7\%. Specifically, the proposed GLDFA-Net surpasses DG-Net \cite{2019DG-NetZheng} with +6.4\% and +3.0\% in mAP and rank-1 accuracy, respectively. Meanwhile, the proposed GLDFA-Net exceeds SFT \cite{2019SFTLuo}  with +0.4\% and +1.2\% in mAP and rank-1 accuracy, respectively. The comparative results validate the superiority of the proposed GLDFA-Net.
\section{Conclusion}\label{Conclusion}
In this paper, we construct a global-local dynamic feature alignment network (GLDFA-Net) for the person Re-ID task. We first propose LSA, a simple and efficient local sliding feature alignment strategy that can dynamically align local features of pedestrians by setting sliding windows on their local stripes. LSA can effectively suppress spatially misalignment and the noises from unshared regions and does not require the introduction of additional auxiliary pose information. We then introduce LSA into the local branch of GLDFA-Net for guiding the computation of distance metrics, which can further improve the performance of the model. Evaluation experiments on the Market-1501, DukeMTMC-reID, CUHK03 and MSMT17 clearly show that the proposed GLDFA-Net has reached the latest experimental performance. Our proposed method has some limitations, such as poor performance in the case of severe pose changes and domain gaps. In the future, we will introduce knowledge distillation and transfer learning technologies to extend GLDFA-Net to cross-domain person Re-ID tasks.

\section*{Declaration of Competing Interest}{The authors declare that they have no known competing financial interests or personal relationships that could have appeared to influence the work reported in this paper.}

\section*{Acknowledgement}{The authors wish to thank Xiao Pang, Jiamin Zhu, Fuqiu Chen, Yi Zhou and Cheng Zhang. This work was supported by the National Key Research and Development Project of China (No. JG2018190), and in part by the National Natural Science Foundation of China (No. 61872256)}

\bibliography{sn-bibliography}


\end{document}